\documentclass{article}

\PassOptionsToPackage{numbers, compress}{natbib}
\usepackage[final]{nips_2016_arxiv}

\usepackage[utf8]{inputenc} 
\usepackage[T1]{fontenc}    
\usepackage{hyperref}       
\usepackage{url}            
\usepackage{booktabs}       
\usepackage{amsfonts}       
\usepackage{nicefrac}       
\usepackage{microtype}      
\usepackage{graphicx}
\usepackage{algorithm}
\usepackage[noend]{algpseudocode}
\usepackage{amsmath}

\usepackage{amsmath,amsthm,amssymb}

\usepackage{todonotes}

\newcommand\cut[1]{}

\newcommand{\squishlist}{
   \begin{list}{$\bullet$}
    { \setlength{\itemsep}{0pt}      \setlength{\parsep}{3pt}
      \setlength{\topsep}{3pt}       \setlength{\partopsep}{0pt}
      \setlength{\leftmargin}{1.5em} \setlength{\labelwidth}{1em}
      \setlength{\labelsep}{0.5em} } }

\newcommand{\squishlisttwo}{
   \begin{list}{$\bullet$}
    { \setlength{\itemsep}{0pt}    \setlength{\parsep}{0pt}
      \setlength{\topsep}{0pt}     \setlength{\partopsep}{0pt}
      \setlength{\leftmargin}{2em} \setlength{\labelwidth}{1.5em}
      \setlength{\labelsep}{0.5em} } }

\newcommand{\squishend}{
    \end{list}  }

\newcommand{\real}{\mathbb{R}}
\newcommand{\myvec}[1]{\mathbf{#1}}
\newcommand{\myvecsym}[1]{\boldsymbol{#1}}

\newcommand{\vphi}{\myvecsym{\phi}}

\newcommand{\vtheta}{\myvecsym{\theta}}

\newcommand{\vv}{\myvec{v}}

\newcommand{\vx}{\myvec{x}}

\newcommand{\vz}{\myvec{z}}
\newcommand{\vzdiff}{\myvec{z}_{diff}}

\newcommand{\be}{\begin{equation}}
\newcommand{\ee}{\end{equation}}
\newcommand{\bea}{\begin{eqnarray}}
\newcommand{\eea}{\end{eqnarray}}
\newcommand{\beaa}{\begin{eqnarray*}}
\newcommand{\eeaa}{\end{eqnarray*}}

\DeclareMathAlphabet{\mathpzc}{OT1}{pzc}{m}{n}

\graphicspath{{figures/}}
\makeatletter
\def\BState{\State\hskip-\ALG@thistlm}
\makeatother

\title{Early Visual Concept Learning \\ with Unsupervised Deep Learning }

\author{
 Irina~Higgins, Loic~Matthey, Xavier~Glorot, Arka~Pal, Benigno~Uria, \\
 \textbf{Charles~Blundell, Shakir~Mohamed, Alexander~Lerchner}\\
 Google DeepMind\\
 \texttt{\{irinah,lmatthey,glorotx,arkap,buria,cblundell,shakir,lerchner\}@google.com}
}

\begin{document}
\maketitle

\begin{abstract}
Automated discovery of early visual concepts from raw image data is a major open challenge in AI research. Addressing this problem, we propose an unsupervised approach for learning disentangled representations of the underlying factors of variation. We draw inspiration from neuroscience, and show how this can be achieved in an unsupervised generative model by applying the same learning pressures as have been suggested to act in the ventral visual stream in the brain. By enforcing redundancy reduction, encouraging statistical independence, and exposure to data with transform continuities analogous to those to which human infants are exposed, we obtain a variational autoencoder (VAE) framework capable of learning disentangled factors. Our approach makes few assumptions and works well across a wide variety of datasets. Furthermore, our solution has useful emergent properties, such as zero-shot inference and an intuitive understanding of ``objectness''.
\end{abstract}

\section{Introduction}
State-of-the-art AI approaches still struggle with some scenarios where humans excel \cite{Lake_etal_2016}, such as knowledge transfer, where faster learning is achieved by reusing learnt representations for numerous tasks (Fig.~\ref{fig_theory}A); or zero-shot inference, where reasoning about new data is enabled by recombining previously learnt factors (Fig.~\ref{fig_theory}B). \cite{Lake_etal_2016} suggest incorporating certain ``start-up'' abilities into deep models, such as intuitive understanding of physics, to help bootstrap learning in these scenarios. Elaborating on this idea, we believe that learning basic visual concepts, such as the ``objectness'' of things in the world, and the ability to reason about objects in terms of the generative factors that specify their properties, is an important step towards building machines that learn and think like people. We believe that this can be achieved by learning a disentangled posterior distribution of the generative factors of the observed sensory input by leveraging the wealth of unsupervised data \cite{Bengio_etal_2013, Lake_etal_2016}. We wish to learn a representation where single latent units are sensitive to changes in single generative factors, while being relatively invariant to changes in other factors \cite{Bengio_etal_2013}. With a disentangled representation, knowledge about one factor could generalise to many configurations of other factors, thus capturing the ``multiple explanatory factors'' and ``shared factors across tasks'' priors suggested by \cite{Bengio_etal_2013}. Unsupervised disentangled factor learning from raw image data is a major open challenge in AI. Most previous attempts require a priori knowledge of the number and/or nature of the data generative factors \cite{Hinton_etal_2011, Reed_etal_2014, Zhu14, Yang_etal_2015, Goroshin_etal_2015, Kulkarni_etal_2015, Cheung15, Whitney_etal_2016, Karaletsos_etal_2016}. This is infeasible in the real world, where the newborn learner may have no a priori knowledge and little to no supervision for discovering the generative factors. So far any purely unsupervised approaches to disentangled factor learning have not scaled well \cite{Desjardins_etal_2012, Tang13, Cohen_Welling_2014, Cohen_Welling_2015}.

We propose a deep unsupervised generative approach for disentangled factor learning inspired by neuroscience \cite{Barlow_1961, Barlow_etal_1989, Perry_etal_2010, Higgins_Stringer_2011}. We apply similar learning constraints to the model as have been suggested to act in the ventral visual stream in the brain \cite{Schenk_McIntosh_2010}: redundancy reduction, an emphasis on learning statistically independent factors, and exposure to data with transform continuities analogous to those human infants are exposed to \cite{Barlow_1961, Barlow_etal_1989}. We show that the application of such pressures to a deep unsupervised generative model can be realised in the variational autoencoder (VAE) framework \cite{Kingma_Welling_2014, Rezende_etal_2014}. Our main contributions are the following: 1) we show the importance of neuroscience inspired constraints (data continuity, redundancy reduction and statistical independence) for learning disentangled representations of continuous visual generative factors; 2) we devise a protocol to quantitatively compare the degree of disentanglement learnt by different models; and 3) we demonstrate how learning disentangled representations enables zero-shot inference and the emergence of basic visual concepts, such as ``objectness''.

\section{Constraints to encourage disentangled factor learning}
\label{sec_theory}
The infants' ventral visual stream learns basic visual concepts through exposure to unsupervised data during the first few months of life \cite{Schenk_McIntosh_2010, Bremner_etal_2015}. We hypothesise that a deep unsupervised model should be able to learn similar representations if exposed to similar data streams and put under the same learning constraints as the visual brain. In this section we elaborate on this hypothesis.

\begin{figure}[t]
 \centering
 \includegraphics[width=0.98\textwidth]{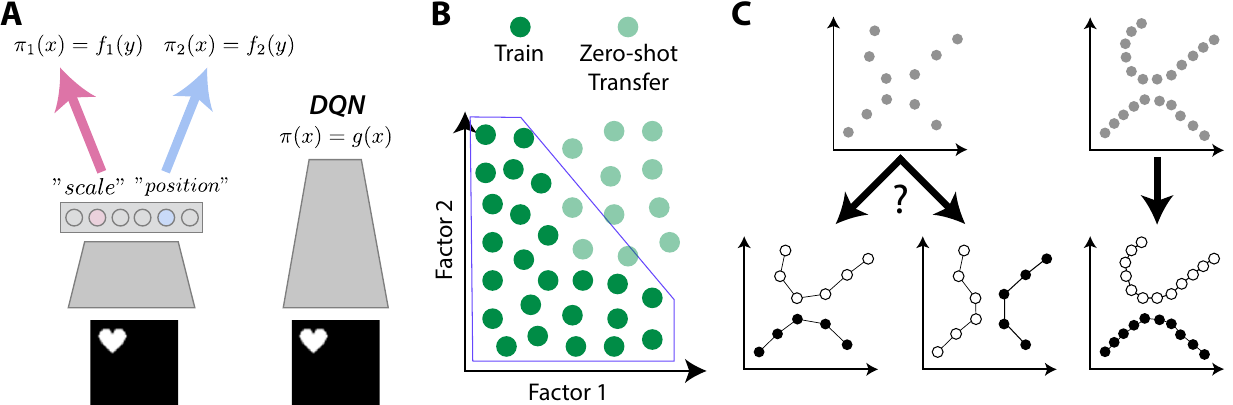}
 \caption{\textbf{A}: Disentangled representations of data generative factors allow for fast knoweldge transfer between different reinforcement learning (RL) policies. State of the art RL models without such representations (e.g. DQN by \cite{Mnih_etal_2015}) require complete re-learning of low-level features for different tasks \cite{Lake_etal_2016}. \textbf{B}: Models are unable to generalise to data outside of the convex hull of the training distribution (light blue line) unless they learn about the data generative factors and recombine them in novel ways. \textbf{C}: Sparse data points do not provide enough information for an unsupervised model to identify where the data manifold should lie. Data generated using factors densely sampled from continuous distributions makes manifold learning less ambiguous.}
 \label{fig_theory}
\end{figure}

\paragraph{Continuously transformed data}
\label{sec_data_stats}
Up to around 3 months of age human babies are unable to focus beyond 8-10 inches  \cite{Leat_etal_2009}. Their visual cortices are learning from a large unsupervised dataset of objects transforming continuously against a blurred background \cite{Candy_etal_2009}. Computational neuroscience simulations of the ventral visual pathway suggest that the response properties of neurons in the inferior temporal cortex arise through a Hebbian learning algorithm that relies on the fact that nearest neighbours of a particular object in pixel space are transforms of the same object \cite{Perry_etal_2010}. This notion can be generalised within the manifold learning framework. As shown in Fig.~\ref{fig_theory}C, sparse samples from data transformation manifolds provide little information for unsupervised models about the manifold shapes. This ambiguity may be resolved through either dense sampling of the manifolds or by adding supervised signals. The importance of vast quantities of unlabeled data for the success of unsupervised approaches in learning disentangled factor representations was pointed by \cite{Bengio_etal_2013}. Here we specify a particular aspect of the data we believe is important for such learning. We postulate that it is important that the \emph{observed data is generated using factors of variation that are densely sampled from their respective continuous distributions}. We leave the learning of discrete factors to future work.

\paragraph{Redundancy reduction and independence}
According to \cite{Barlow_1961}, one of the main functions of the sensory brain is redundancy reduction, where redundacy is defined as the difference between the maximum entropy that a channel can transmit, and the entropy of messages actually transmitted. Sensory redundancy reduction is facilitated through learning statistically independent components within the data \cite{Barlow_etal_1989}. We hypothesise that an unsupervised deep model encouraged to perform redundancy reduction and to learn statistically independent components from continuous data, as described in the section above, will learn basic visual concepts similar to those learnt by the ventral visual stream. Such constraints have been considered before \cite{Vasilescu_Terzopoulos_2005, Schmidhuber_1992, Rippel_Adams_2013}, but no scalable unsupervised solution capable of disentangled factor learning based on these ideas yet exists.

We start by specifying an unsupervised deep generative model for learning latent factors $\vz \in \real^m$ that, when combined in a non-linear way, generate the observed data $\vx$. For a given observation, we describe the plausible posterior configurations of such generative latent factors $\vz$ by a probability distribution $q_\phi(\vz | \vx)$. We aim to maximise the probability of the observed data $\vx$ on average over all possible samples from the latent factors $\vz$. This corresponds to the optimisation problem in Eq.~\ref{eq_1}.

\vspace{-10pt}
\begin{equation}
\label{eq_1}
\max_{\phi, \theta} \mathbb{E}_{q_\phi(\vz|\vx)} [\log p_\theta(\vx |\vz)]
\end{equation}
\vspace{-10pt}

In order to learn disentangled representations of the generative factors we introduce a constraint that encourages the distribution over latent factors $\vz$ to be close to a prior that embodies the neuroscience inspired pressures of redundancy reduction and independence prior. This results in a constrained optimisation problem shown in Eq.~\ref{eq_2}, where $\epsilon$ specifies the strength of the applied constraint.

\vspace{-10pt}
\begin{equation}
\label{eq_2}
\max_{\phi, \theta} \mathbb{E}_{q_\phi(\vz|\vx)} [\log p_\theta(\vx |\vz)] \quad \textrm{ subject to } \ D_{KL}(q_\phi(\vz|\vx) || p(\vz)) < \epsilon.
\end{equation}
\vspace{-10pt}

Writing Eq.~\ref{eq_2} as a Lagrangian we obtain the familiar variational free energy objective function shown in Eq.~\ref{eq_3} \cite{Kingma_Welling_2014, Rezende_etal_2014}, where $\beta \geqslant 0$ is the inverse temperature or regularisation coefficient.

\vspace{-10pt}
\begin{equation}
\label{eq_3}
	\mathcal{L}(\vtheta, \vphi; \vx)  =  \mathbb{E}_{q_\phi(\vz|\vx)} [\log p_\theta(\vx |\vz)] - \beta \ D_{KL}(q_\phi(\vz|\vx) || p(\vz))
\end{equation}
\vspace{-10pt}

If we set the disentangled prior to be an isotropic unit Gaussian ($p(\vz)=\mathcal{N}(0,I)$), the variational bound in Eq.~\ref{eq_3} matches well the desiderata proposed by \cite{Barlow_1961, Barlow_etal_1989}. It adds redundancy reduction pressure by constraining the capacity of the latent information channel $\vz$, while preserving enough information to enable reconstruction. The isotropic nature of the Gaussian prior puts implicit independence pressure on the learnt posterior. Varying $\beta$ changes the degree of applied learning pressure during training, thus encouraging different learnt representations. When $\beta = 0$, we obtain the standard maximum likelihood learning. When $\beta = 1$, we recover the Bayes solution. We postulate that in order to learn disentangled representations of the continuous data generative factors it is important to \emph{tune $\beta$ to approximate the level of learning pressure present in the ventral visual stream}.

\section{Experiments}
\subsection{Learning disentangled factors in a 2D dataset}
\label{sec_2D}
\begin{figure}[t]
 \centering
 \includegraphics[width = 0.99\textwidth]{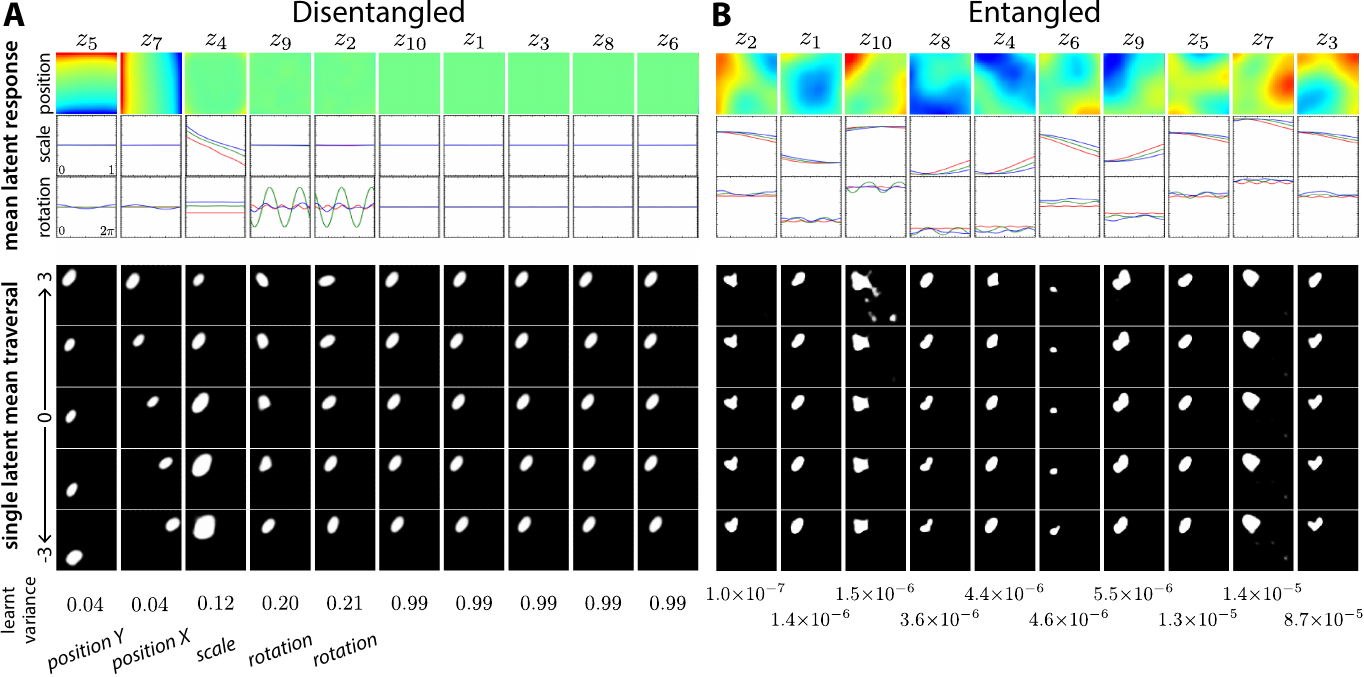}
 \caption{\textbf{A:} Disentangled representation learnt with $\beta=4$. Each column represents a latent $z_i$, ordered according to the learnt Gaussian variance (last row). Row 1 (position) shows the mean activation (red represents high values) of each latent $z_i$ as a function of all 32x32 locations averaged across objects, rotations and scales. Row 2 (scale) shows the mean activation of each unit $z_i$ as a function of scale (averaged across rotations and positions). Row 3 (rotation) shows the mean activation of each unit $z_i$ as a function of rotation (averaged across scales and positions). \emph{Square} is red, \emph{oval} is green and \emph{heart} is blue. Rows 4-8 (second group) show reconstructions resulting from the traversal of each latent $z_i$ over three standard deviations around the unit Gaussian prior mean while keeping the remaining 9/10 latent units fixed to the values obtained by running inference on an image from the dataset. After learning, five latents learnt to represent the generative factors of the data, while the others converged to the uninformative unit Gaussian prior. \textbf{B:} Similar analysis for an entangled representation learnt with $\beta=0$.
 }
 \label{fig_2D_latents_analysis}
\end{figure}

We first demonstrate that a VAE can learn disentangled generative factors when exposed to a dataset with continuous transformations as defined in Sec.~\ref{sec_theory}. We use a synthetic binary dataset of 737,280 2D shapes (heart, oval and square) generated from the Cartesian product of four factor values $v_k$ defined in vector graphics: position X (32 values), position Y (32 values), scale (6 values) and rotation (40 values over the $2\pi$ range). To ensure smooth affine object transforms, each two subsequent values for each factor $v_k$ were chosen to ensure minimal differences in pixel space given 64x64 pixel image resolution. We used randomly sampled batches of size 100 to train a fully connected VAE with $m=10$ latent units and various $\beta$ values until convergence (see Tbl.~\ref{tbl_models} in Appendix for details). After training, a VAE with $\beta=4$ learnt a good (while not perfect) disentangled representation of the data generative factors, and its decoder learnt to act as a rendering engine (Fig.~\ref{fig_2D_latents_analysis}A). The most informative units $z_i$ have the highest KL divergence from the unit Gaussian prior ($p(z)=\mathcal{N}(0,I)$), while the uninformative latents have KL divergence close to zero. Throughout the rest of the paper we illustrate disentangling performance of various models using latents with the highest KL divergence from the prior.

Fig.~\ref{fig_2D_latents_analysis}A demonstrates the selectivity of each latent $z_i$ to the continuous data generating factors: $z_i^\mu = f(v_k) ~\forall v_k \in \{v_{positionX}, v_{positionY}, v_{scale}, v_{rotation}\}$ (top three rows), where $z_i^\mu$ stands for the learnt Gaussian mean of latent unit $z_i$. The effect of traversing each latent $z_i$ on the resulting reconstructions is shown in the bottom five rows of Fig.~\ref{fig_2D_latents_analysis}A. It can be seen that latents $z_7$ and $z_5$ learnt to encode X and Y coordinates of the objects respectively; unit $z_4$ learnt to encode scale; and units $z_2$ and $z_9$ learnt to encode rotation. The frequency of oscilations in each rotational latent corresponds to the rotational symmetry of the corresponding object ($2\pi$ for heart, $\pi$ for oval and $\pi/2$ for square). Furthermore, the two rotational latents seem to encode $\cos$ and $\sin$ rotational coordinates, while the positional latents align with the Cartesian axes. While such alignment with human intuition is not guaranteed, empirically we found it very common. Fig.~\ref{fig_2D_latents_analysis}B demonstrates that a model with inappropriate learning pressures ($\beta=0$) does not learn about the generative factors in the data and instead learns a dense entangled latent representation.

\subsection{Quantifying disentangling}
\label{sec_quant}
\begin{figure}[t]
 \centering
 \includegraphics[width = 0.99\textwidth]{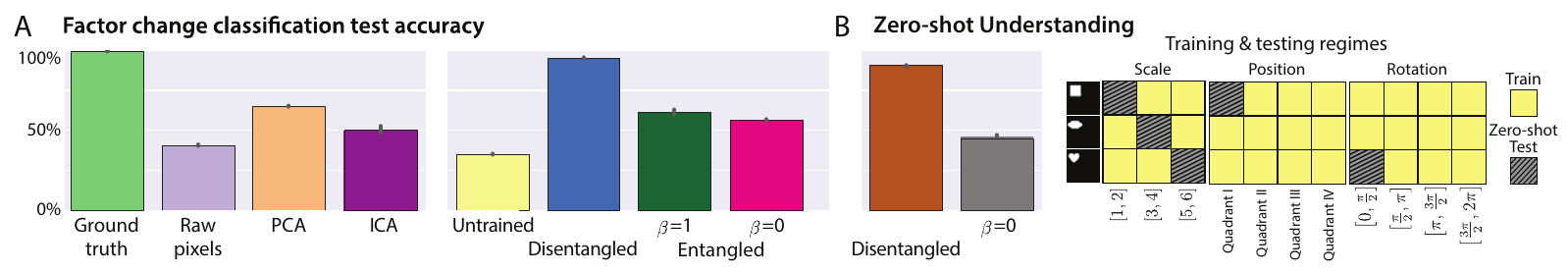}
 \caption{\textbf{A:} Factor change classification accuracy for the original 2D shapes dataset (heart, oval and square). Ground truth uses data generating vectors $v$.  PCA and ICA decompositions keep the first ten components (PCA components explain $60.8\%$ of variance). Untrained refers to a VAE with random weights. Disentangled is a VAE with $\beta=4$. Entangled uses either $\beta=0$ (maximum likelihood solution) or $\beta=1$ (Bayes solution). \textbf{B:} ``Zero-shot Understanding'' refers to a VAE that did not see particular combinations of the generative factors during training (see Sec.~\ref{sec_zsl}), but had to reason about them during factor change classification. A projection of the hypercube formed by the data generative factors is visualised on the right. Only the yellow subset was used for training. The held out factor combinations are shown in grey and were used to evaluate the factor change classification accuracy.}
 \label{fig_main_results}
\end{figure}

We have devised a metric to quantitatively approximate the degree of disentanglement within the learnt latent representations. The metric uses a linear classifier to predict which factor caused the transition between two frames in the dataset, where the frames are identical apart from a random change in a single generative factor. We use a low VC dimension classifier that has no capacity to do the disentangling itself to ensure that good classification performance can be achieved only if the generative factors are already disentangled in the latent space $\vz$. The classifier has to learn a mapping  $G(\vzdiff): \mathcal{R}^m\rightarrow \mathcal{R}^k$, where $m$ is the dimensionality of the latent space $\vz$, $k$ is the number of factors in the dataset (in our case four: scale, rotation, position X and position Y) and $\vzdiff = \frac{|\vz_{start}^{\mu} - \vz_{end}^{\mu}|}{max(|\vz_{start}^{\mu} - \vz_{end}^{\mu}|)}$ is the change in the latent space corresponding to a change in a single generative factor in pixel space (see Alg.~1 in Appendix for details). Classification performance is reported for 5,000 test samples in Fig.~\ref{fig_main_results}. VAE that learnt a disentangled representation of the data generating factors (model in Fig.~\ref{fig_2D_latents_analysis}A) achieved similar classification score to the one obtained using ground truth data generation vectors $\vv_{diff}$. Both scores are significantly higher than several varied baselines: an untrained VAE with the same architecture, a VAE that matches the Bayes solution ($\beta=1$), a VAE that matches the maximum likelihood solution ($\beta=0$, model in Fig.~\ref{fig_2D_latents_analysis}B), the top ten PCA ($PCA_{diff}$) or ICA ($ICA_{diff}$) components of the data or using the raw pixels ($\vx_{diff}$), see Fig.~\ref{fig_main_results}A.

\subsection{Factors affecting learning}
In this section we investigate the sensitivity of disentangled factor learning in the VAE framework to the learning constraints of data continuity, redundancy reduction and independence.

\begin{figure}[t]
 \centering
 \includegraphics[width = 0.99\textwidth]{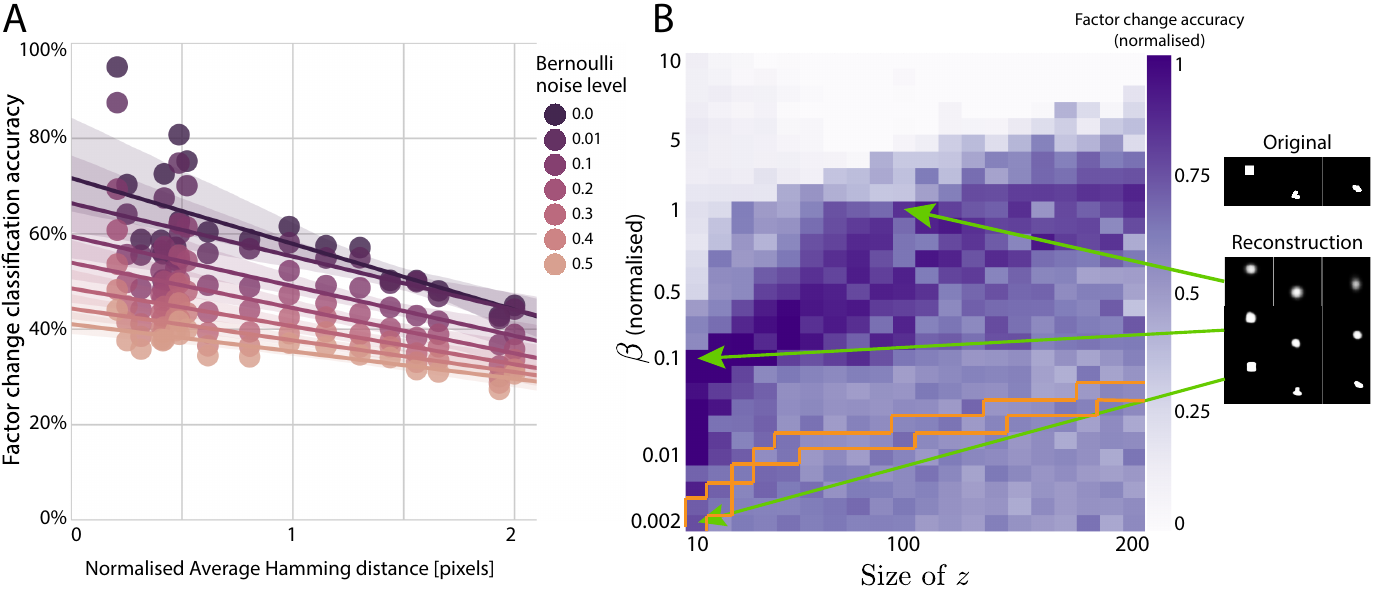}
 \vspace{-6pt}
 \caption{\textbf{A:} Negative correlation between data transform continuity and the degree of disentangling achieved by VAEs. Abscissa is the average normalized Hamming distance between each of the two consecutive transforms of each object. Ordinate is factor change classification accuracy from Sec.~\ref{sec_quant}. Disentangling performance is robust to Bernoulli noise added to the data at test time, as shown by slowly degrading classification accuracy up to 10\% noise level, considering that the 2D objects occupy on average between 2-7\% of the image depending on scale. Fluctuations in classification accuracy for similar Hamming distances are due the different nature of subsampled generative factors (i.e. symmetries are present in rotation but are lacking in position). \textbf{B: } Positive correlation is present between the size of $\vz$ and the optimal \emph{normalised} values of $\beta$ for disentangled factor learning for a fixed VAE architecture. $\beta$ values are normalised by latent $\vz$ size $m$ and input $\vx$ size $n$. Note that $\beta$ values are not uniformly sampled. Good reconstructions are associated with entangled representations (lower disentanglement scores). Orange approximately corresponds to \emph{unnormalised} $\beta=1$. Disentangled representations (high disentanglement scores) often result in blurry reconstructions.}
 \label{fig_cont_kl}
\end{figure}

\paragraph{Data continuity}
\label{sec_data_cont}
We hypothesised that data continuity is important for guiding unsupervised models towards learning the correct data manifolds (Sec.~\ref{sec_theory}). To test this idea we measured how the degree of learnt disentangling changes with reduced continuity in the 2D shapes dataset. We trained a VAE with $\beta=4$ (Fig.~\ref{fig_2D_latents_analysis}A) on subsamples of the original 2D shapes dataset, where we progressively  decreased the generative factor sampling density. Reduction in data continuity negatively correlates with the average pixel wise (Hamming) distance between two consecutive transforms of each object (normalised by the average number of pixels occupied by each of the two adjacent transforms of an object to account for object scale). Fig.~\ref{fig_cont_kl}A demonstrates that as the continuity in the data reduces, the degree of disentanglement in the learnt representations also drops. This effect holds after additional hyperparameter tuning and can not solely be explained by the decrease in dataset size, since the same VAE can learn disentangled representations from a data subset that preserves data continuity but is approximately 55\% of the original size (see Sec.~\ref{sec_zsl}).

\paragraph{Optimizing learning constraints}
\label{sec_kl}
We hypothesised that constrained optimisation is important for enabling deep unsupervised models to learn disentangled representations of the data generative factors (Sec.~\ref{sec_theory}). In the VAE framework this corresponds to tuning the $\beta$ coefficient. One way to view $\beta$ is as a mixing coefficient for balancing the magnitudes of gradients from the reconstruction and the prior-matching costs when training the VAE encoder. In this context it makes sense to normalise $\beta$ by latent $\vz$ size $m$ and input $\vx$ size $n$ in order to compare its different values across different latent layer sizes. It can be seen that larger latent $\vz$ layer sizes $m$ require higher constraint pressures (higher normalised $\beta$ values) (Fig.~\ref{fig_cont_kl}B). Furthermore, the relationship of $\beta$ for a given $m$ is characterised by an inverted U curve. When $\beta$ is too low or too high the model learns an entangled latent representation due to either too much or too little capacity in the latent $\vz$ bottleneck. We find that in general unnormalised $\beta>1$ is necessary to achieve good disentanglement. We also note that VAE reconstruction quality is a poor indicator of learnt disentanglement. Good disentangled representations often lead to blurry reconstructions due to restricted capacity of the latent information channel $\vz$, while entangled representations often result in the sharpest reconstructions. Since VAE model selection is often performed based on reconstruction quality, this may be one of the reasons why the ability of VAEs to disentangle data generative factors has been overlooked before. Another reason may be the lack of transform continuity in many popular datasets (i. e. Multi-PIE \cite{Gross_etal_2010}).

\subsection{Investigating qualities of learnt representations}
In this section we show some of the desirable properties that arise from learning disentangled as opposed to entangled latent representations.

\begin{figure}[t]
 \centering
 \includegraphics[scale = 1]{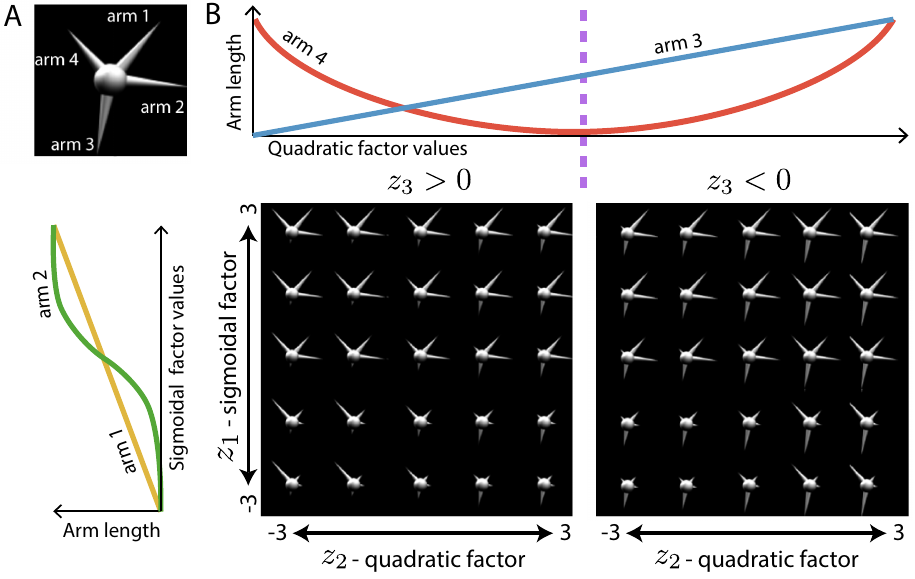}
 \caption{\textbf{A}: Amoeba object with four arms of varying length. \textbf{B}: Two non-linear generative factors determine the lengths of the pairwise grouped arms. Traversal over three standard deviations around the unit Gaussian prior mean for two latent units ($z_1$ and $z_2$) that learnt disentangled representations of the two generative factors. $z_1$ learnt the sigmoidal factor. $z_2$ learnt the quadratic factor. $z_3$ learnt to be a switch that determines which half of the quadratic factor is traversed by $z_2$.}
 \label{fig_amoeba}
\end{figure}

\paragraph{Learning statistically independent factors}
\label{sec_amoeba}
Computational neuroscience results suggest that the nature of representations learnt through Hebbian learning in the ventral visual stream in the brain relies on the statistics of the data.  Statistically independent parts of the retinal inputs are allocated separate representations, while statistically dependent parts are grouped into a single representation \cite{Higgins_Stringer_2011}. We test whether the same holds for VAEs trained for disentangled factor learning. We use a dataset developed for psychophysical experiments to measure generative factor learning in humans \cite{Chadwick_Kumaran_2015} (unpublished). The dataset consists of a single ``amoeba'' object with four arms of varying length (Fig.~\ref{fig_amoeba}A). The arms are pairwise coupled and the length of each arm within each pair is determined by a nonlinear factor (either quadratic or sigmoidal, see Fig.~\ref{fig_amoeba}B). For example, growth in the values of the quadratic factor correspond to linear growth of arm three and quadratic growth of arm four. This means that during training the VAE sees the full range of lengths of each single arm, but it never sees certain combinations of lengths of pairs of arms (i.e. long arm three and a short arm four). We investigated whether a fully connected VAE (see Tbl~\ref{tbl_models} for architecture details) would learn representations of the two generative factors (sigmoidal and quadratic), or whether it would learn four separate representations, one for each arm (the latter would be expected if the VAE did not learn the statistical regularities in the data). We found the former to be true (Fig.~\ref{fig_amoeba}B, $\beta=16.38$): the VAE learnt to allocate two latents ($z_1$ and $z_2$) to represent the sigmoidal and quadratic factors respectively, $z_3$ acted as a switch to split the quadratic factor space into two halves, while the remaining latents ($z_4$-$z_{10}$) learnt the uninformative unit Gaussian prior ($p(z)=\mathcal{N}(0,I)$).

\paragraph{Generalisation to new latent factor combinations}
\label{sec_zsl}
A model that understands the factorial structure of the data should be able to generalize its knowledge beyond the training distribution by recombining previously learnt factor values, thus performing zero-shot inference (Fig.~\ref{fig_theory}B). We tested such properties of VAEs by training the architecture described in Sec.~\ref{sec_2D} on a subset of the full 2D shapes dataset. This subset preserved the original data continuity by traversing each individual generative factor fully, but some combinations of factors were never seen during training (i.e. the subset still contained all six scales across the three object identities, but there were no small squares present in any rotation or position). By dropping certain combinations of generative factors (Fig.~\ref{fig_main_results}B) we reduced the dataset size to approximately 55\% of the original size. We then calculated the disentangling metric (Sec.~\ref{sec_quant}) for a model with $\beta=4$ or $\beta=0$ trained on this subset. The disentangling metric was calculated using factor combinations that were excluded from the training subset. We found that the VAE with $\beta=4$ learnt a disentangled representation and was able to reason well about the test data significantly outside of its training distribution (Fig.~\ref{fig_main_results}B). The model with $\beta=0$ learnt an entangled representation and had significantly worse generalization to test data outside of the convex hull of its training data distribution.

\begin{figure}[t]
 \centering
 \includegraphics[scale = 0.9]{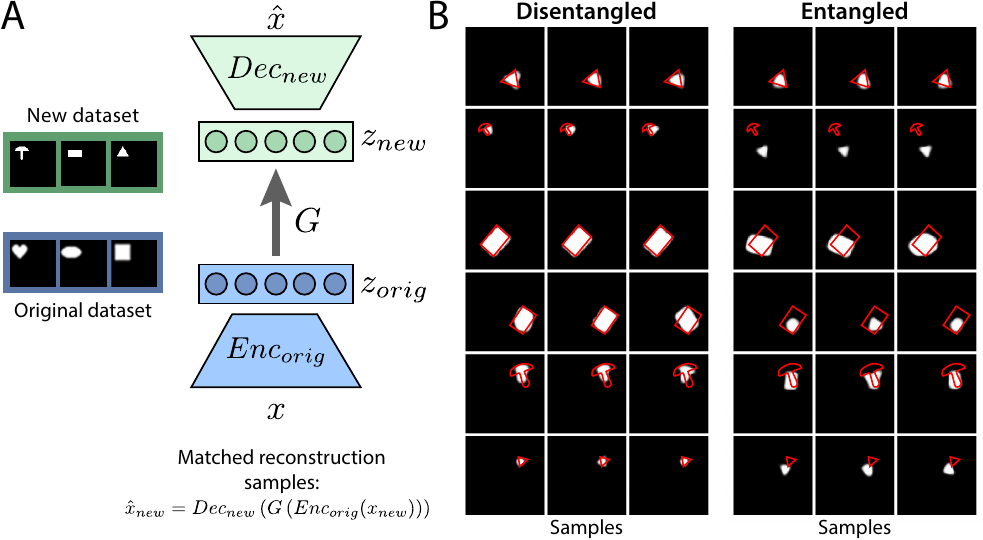}
 \caption{\textbf{A}: Model architecture used to visualise whether VAEs trained on the original dataset of 2D objects (\emph{heart}, \emph{oval} and \emph{square}) can reason about new object identities (\emph{mushroom}, \emph{rectangle} and \emph{triangle}). We splice an encoder trained on the original dataset ($Enc_{orig}$) with a decoder trained on the new dataset ($Dec_{new}$) using a linear regressor $G$, which learns to align the latent spaces $\vz_{orig}$ and $\vz_{new}$. \textbf{B}: Samples from $G(z_{orig})$ when running inference through $Enc_{orig}$ using novel 2D objects. Each row corresponds to a different ground truth image $x_{new}$ (red outline). Disentangled VAE reasons well about the location, position and rotation of the novel objects, while slightly confusing object identities; the average normalized Hamming distance between original and reconstructed images over the whole new dataset is $0.42$. Entangled VAE struggles to reason about the new objects. Its average normalized Hamming distance is $0.93$.}
 \label{fig_newdata}
\end{figure}

\paragraph{Learning basic visual concepts}
\label{sec_robust}
We argue that through learning disentangled representations of the data generative factors, VAEs may acquire basic conceptual understanding of the visual world, such as the ``objectness'' of things. Then, when presented with novel objects, the VAEs may still be able to reason about the properties of these objects, such as size or position, without necessarily knowing the identify of the new objects. A reinforcement learning framework built on top of such a VAE will then be able to preserve its policy performance without re-learning, hence moving towards the desiderata described in \cite{Lake_etal_2016}. To test this we presented models that learnt disentangled ($\beta=4$) or entangled ($\beta=0$) representations on the original dataset of 2D objects (heart, oval and square) with new 2D objects (mushroom, rectangle and triangle) generated using the same four factors of variation (scale, rotation, position X and position Y). In order to visualise what exactly the VAE understands about the new 2D objects, we spliced together an encoder trained on the original dataset ($Enc_{orig}=p(z_{orig}|x_{orig})$) with a decoder trained on the new 2D shapes dataset ($Dec_{new}=p(x_{new}|z_{new})$) (Fig.~\ref{fig_newdata}A). We used a low VC dimension linear regressor to learn an alignment mapping $G: z_{orig}\rightarrow z_{new}$ using 50\% of the new dataset. We then generated reconstructions $\hat{x}_{new} = Dec_{new}(G(Enc_{orig}(x_{new})))$ of the held out test data. Fig.~\ref{fig_newdata}B shows that the VAE that learnt a disentangled representation can reason well about location, scale, and rotation of the new objects despite the fact that its encoder has never seen the new objects. This is in contrast to the poor reconstructions produced by a VAE with an entangled representation.

\subsection{Other datasets}
\label{sec_other}
\begin{figure}[t]
 \centering
 \includegraphics[width = 0.99\textwidth]{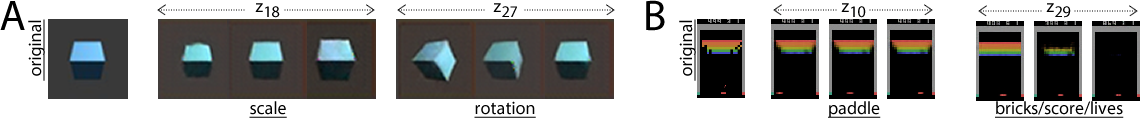}
 \caption{Best seen in animation at \url{http://tinyurl.com/jgbyzke}. Examples of disentangled factors learnt for different datasets. We run inference on an original image from each dataset, clamp all latent units to the values obtained, then traverse units $z_i$ one at a time. Reconstructions shown are generated by traversing $z_i$’s with the lowest learnt prior variance for each dataset. \textbf{A}: synthetic dataset of 3D shapes with non-affine transformations. \textbf{B}: Atari game Breakout.}
 \label{fig_other}
\end{figure}

Additionally, we trained convolutional VAE architectures on a variety of datasets (including a 3D first person view maze navigation environment that shares many properties with the real world) and found them to robustly learn disentangled generative factor representations (see Tbl.~\ref{tbl_models} for details of VAE architectures and datasets). Some examples of learnt disentangled factors are shown in Fig.~\ref{fig_other}, however these are best seen in animations at \url{http://tinyurl.com/jgbyzke}. The examples of learnt factors include non-affine rotation of 3D shapes ($m=10$, $\beta=1$) the movement of the paddle or changing the score in the Atari game Breakout ($m=30$, $\beta=1.28$); the forward/rotational movement in a 3D first person view maze navigation environment ($m=32$, $\beta=1$), and the rotation of chairs in a dataset of 3D chairs \cite{Aubry_etal_2014} ($m=10$, $\beta=1$). Equivalent architectures that lacked the learning pressures necessary for disentangled factor learning ($\beta=0$) could not disentangle the latent factors (results in \url{http://tinyurl.com/jgbyzke}).

\section{Conclusion}
In this paper we have shown that deep unsupervised generative models are capable of learning disentangled representations of the visual data generative factors if put under similar learning constraints as those present in the ventral visual pathway in the brain: 1) the observed data is generated by underlying factors that are densely sampled from their respective continuous distributions; and 2) the model is encouraged to perform redundancy reduction and to pay attention to statistical independencies in the observed data. The application of such pressures to an unsupervised generative model leads to the familiar VAE formulation \cite{Kingma_Welling_2014, Rezende_etal_2014} with a temperature coefficient $\beta$ that regulates the strength of such pressures and, as a consequence, the qualitative nature of the representations learnt by the model. Our approach does not depend on any a priori knowledge about the number or the nature of data generative factors, it is robust with respect to different VAE architectures, optimisation parameters, datasets and noise. We have shown that learning disentangled representations leads to useful emergent properties. The ability of trained VAEs to reason about new unseen objects suggests that they have learnt from raw pixels and in a completely unsupervised manner basic visual concepts, such as the ``objectness'' property of the world. This is an important ability for the development of artificial intelligence that understands the world the same way humans do \cite{Lake_etal_2016}. Furthermore, we have demonstrated the ability of VAEs trained for disentangled factor learning to generalise beyond the training data distribution in zero-shot inference scenarios. These are just the first demonstrations of how learning better representations in an unsupervised manner allows models to perform better on challenging machine learning tasks. We believe that using our approach as an unsupervised pre-training stage for supervised or reinforcement learning will produce significant improvements for scenarios such as transfer or fast learning.

\bibliography{disentangledFactors_nips2016}
\bibliographystyle{abbrv} 

\appendix











\section{Appendix}
\label{sup_models}
A summary of all VAE architectures used in this paper can be seen in Tbl~\ref{tbl_models}. Next we provide various auxiliary details for the different datasets.

\begin{table}[h]
 \centering
 \small
 \begin{tabular}{llll}
   \toprule
   Dataset   & Optimiser  & \multicolumn{2}{c}{VAE architecture}    \\
   \midrule
   2D shapes & adagrad \cite{Duchi_etal_2011} & encoder & fc 4096-1200-1200-10 (ReLU)   \\
   & & decoder &  fc 10-1200-1200-1200-4096 (tanh) \\

   3D shapes  & adam \cite{Kingma_Ba_2014} & encoder  &  conv 32x6x6 (2-1)-64x6x6 (2-1)-512-32 (tanh)    \\
   & & decoder & deconv 32-512-32x4x4 (2-1)-64x4x4 (2-1)-128x4x4 (2-1)  \\

   Amoeba  & adagrad \cite{Duchi_etal_2011}  & encoder  &  fc 16384-400-205-10 (ReLU)    \\
   & & decoder & fc 10-400-8392-16384 (ReLU) \\

   Atari (breakout) & adagrad \cite{Duchi_etal_2011} & encoder  &  conv 3x48x80-64x6x6 (2)-32x6x6 (2)-32x5x5 (2)-30  (tanh)   \\
   & & decoder & deconv 30-3840-SU(2)-64x5x5-SU(2)-64x5x5-SU(2)-3x5x5-3x48x80 (tanh) \\

   Atari (other)  & adam \cite{Kingma_Ba_2014}  & encoder  &  conv 32x6x6 (2)-64x6x6 (2-1)-64x6x6 (2-1)-512-various  (ReLU)   \\
   & & decoder & deconv reverse of encoder (ReLU) \\

   3D chairs \cite{Aubry_etal_2014} & rmsprop \cite{Tieleman_Hinton_2012} & encoder & conv 32x6x6 (2)-64x6x6 (2)-256-10 (ReLU) \\
   & & decoder & deconv reverse of encoder (ReLU) \\

   3D game  & rmsprop \cite{Tieleman_Hinton_2012} & encoder  &   conv 3x64x64-32x4x4 (2)-32x5x5 (2)-64x5x5 (2)-64x4x4 (ReLU)  \\
   & & decoder & deconv reverse of encoder (ReLU) \\

   \bottomrule
 \end{tabular}
   \caption{Various VAE architectures and optimisers were used for different experiments to show robustness of our approach. For convolutional architectures the numbers in parenthesis indicate: (stride-padding). SU stands for spatial upsampling.}
   \label{tbl_models}
\end{table}

\subsection{2D shapes dataset}
We trained the fully connected architecture in Tbl.~\ref{tbl_models} with cross-entropy cost function using adagrad \cite{Duchi_etal_2011} with learning rate of 1e-2.

\subsection{Factor change classification}
In order to quantify the degree of disentanglement learnt by the models we generated factor change data according to the pseudocode shown in Algorithm~\ref{alg_factor_change}. We used a linear classifier to learn the identity of the generative factor that produced $z_{diff}$. We used a fully connected neural network mapping between input of size $z_{diff}$ to an output of size 4 corresponding to the 4 generative factors (position X, position Y, scale and rotation) with softmax output nonlinearity and cross-entropy cost function. The classifier was trained with adagrad \cite{Duchi_etal_2011} with learning rate of 1e-2 until convergence.

All factor change classification results reported in the paper were calculated in the following manner. Ten replicas of each VAE experiment was run, each with a different random seed. Each of the ten replicas was evaluated three times using the factor change classification algorithm, each time with a different random seed. We then discarded the bottom 50\% of the thirty resulting scores and reported the remaining results.

\begin{algorithm}
\caption{Data generation for factor change quantification}\label{alg_factor_change}
\begin{algorithmic}[1]
\Procedure{SampleBatch}{}
\For {$n = 1, \textit{batch size} $}
\State $objId \gets \text{randomly sample object identity}$
\State $changeFactor \gets \text{randomly sample factor identity}$
\State $changeDir \gets \text{Randomly sample the direction of change (+/-)}$

\For {$factor \in \{scale, rotation, position X, position Y)\}$}
\State $groundTruth_{start}^{factor} \gets \text{randomly sample factor value}$
\EndFor

\State $groundTruth_{end} \gets groundTruth_{start}$
\State $groundTruth_{end}^{changeFactor} \gets \text{randomly sample a new value in the direction of } changeDir$

\State $x_{start} \gets \text{pixel representation of } groundTruth_{start}$
\State $x_{end} \gets \text{pixel representation of } groundTruth_{end}$

\State $z_{start} \gets Enc(x_{start})$
\State $z_{end} \gets Enc(x_{end})$

\State $z_{diff}^n \gets \frac{|z_{start}^{\mu} - z_{end}^{\mu}|}{max(|z_{start}^{\mu} - z_{end}^{\mu}|)}$
\EndFor
\EndProcedure
\end{algorithmic}
\end{algorithm}

\subsection{Zero shot inference regression}
In order to map $z_{orig}$ to $z_{new}$, we used a fully connected linear neural network with smooth L1 loss trained with adagrad \cite{Duchi_etal_2011} with learning rate of 1e-2 until convergence.

\subsection{Amoeba dataset}
We trained the fully connected architecture in Tbl.~\ref{tbl_models} with binary cross-entropy criterion and adagrad \cite{Duchi_etal_2011} optimizer with learning rate of 1e-2.

\subsection{3D shapes dataset}
We trained a convolutional VAE (see Tbl.~\ref{tbl_models}) with learning rate 1e-4 on a dataset of three 3D objects (cylinder, cube and pyramid) with three factors of variation (6 scales, 60 out of plane rotations and 26 colours). The 3D objects were rotating around the z-axis over $2\pi$ using 60 equidistant steps. The objects were generated in Blender and the 6 scales and 6x6 position translations were generated for each object in each rotational position using ImageMagick. The full dataset contained 38,880 frames of size 64x64. The decoder had Gaussian outputs.

\subsection{Atari dataset}
We trained a convolutional VAE (see Tbl.~\ref{tbl_models}) with learning rate 1e-4 on frames from the Atari games Breakout ($\vz$ size 30, $\beta=1$), SeaQuest ($\vz$ size 10, $\beta=5$), Frostbite ($\vz$ size 100, $\beta=5$) and Enduro ($\vz$ size 100, $\beta=1.75$). The Atari dataset consisted of 1 million frames collected from a trained DQN agent \cite{Mnih_etal_2015}. The frames were pre-processed as described in \cite{Mnih_etal_2015}. The continuity of the dataset enabled the VAE to learn disentangled representations of the independent factors in the data (see video visualisations at http://tinyurl.com/jgbyzke). The decoder had Gaussian outputs. The model was trained using adam \cite{Kingma_Ba_2014} optimizer with learning rate of 1e-4.

\subsection{3D chairs dataset}
For the 3D chairs dataset \cite{Aubry_etal_2014} we trained a convolutional VAE (see Tbl.~\ref{tbl_models}) on 82 chair identities. The images were cropped and downsampled to 100x100 pixels. The decoder had Gaussian outputs. The model was trained using rmsprop \cite{Tieleman_Hinton_2012} optimizer with learning rate of 1e-5.

\subsection{3D first person view maze navigation game dataset}
We also trained a convolutional VAE (see Tbl.~\ref{tbl_models}) on frames from a first person view 3D first person maze navigation game environment. The game frames were made greyscale and downsampled to 84x84 pixels. The dataset contained 1 million frames. This environment shares many properties with the real world: it is continuous and the dynamics of visual scene changes are similar to those experienced in the real world. After training the VAE was able to learn disentangled representations of several factors of variation present in the 3D game world (see video visualisations at http://tinyurl.com/jgbyzke). For example, certain single latent units learnt to represent changes in light, forward/backward movement and rotational movement. The VAE also learnt to allocate single latent units to represent the change in score and the rotation of the little character head at the bottom of the screen. For this experiment $z$ size was set to 32 and $\beta=1$. The decoder had Gaussian outputs. The model was trained using rmsprop \cite{Tieleman_Hinton_2012} optimizer with learning rate of 1e-4.



\end{document}